\pdfoutput=1

\documentclass[11pt]{article}

\usepackage[review]{acl}

\usepackage{times}
\usepackage{latexsym}

\usepackage[T1]{fontenc}

\usepackage[utf8]{inputenc}

\usepackage{microtype}

\title{VGNMN: Video-grounded Neural Module Network to Video-Grounded Language Tasks}

\author{First Author \\
  Affiliation / Address line 1 \\
  Affiliation / Address line 2 \\
  Affiliation / Address line 3 \\
  \texttt{email@domain} \\\And
  Second Author \\
  Affiliation / Address line 1 \\
  Affiliation / Address line 2 \\
  Affiliation / Address line 3 \\
  \texttt{email@domain} \\}

\begin{document}
\maketitle
\begin{abstract}
Neural module networks (NMN) have achieved success in image-grounded tasks such as Visual Question Answering (VQA) on synthetic images. However, very limited work on NMN has been studied in the video-grounded language tasks. These tasks extend the complexity of traditional visual tasks with the additional visual temporal variance. Motivated by recent NMN approaches on image-grounded tasks, we introduce Video-grounded Neural Module Network (VGNMN) to model the information retrieval process in video-grounded language tasks as a pipeline of neural modules. VGNMN first decomposes all language components to explicitly resolve any entity references and detect corresponding action-based inputs from the question. The detected entities and actions are used as parameters to instantiate neural module networks and extract visual cues from the video. Our experiments show that VGNMN can achieve promising performance on two video-grounded language tasks: video QA and video-grounded dialogues. 
\end{abstract}

\section{Introduction}
\begin{figure*}[htbp]
	\centering
	\resizebox{0.9\textwidth}{!} {
	\includegraphics{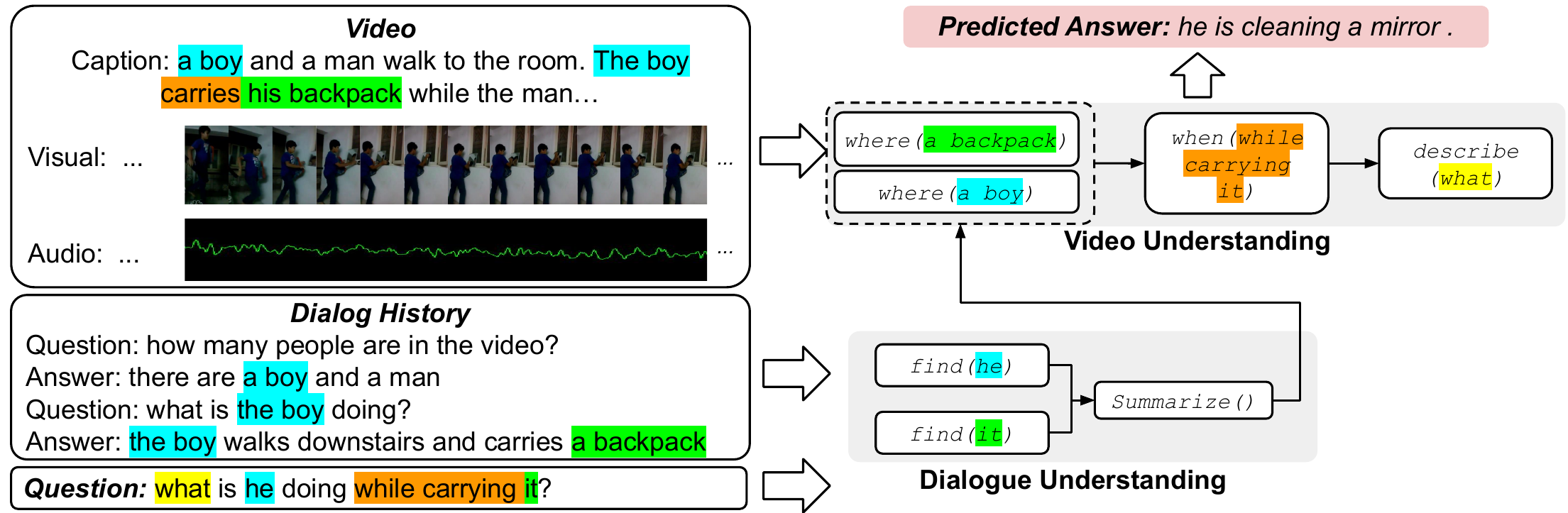}
	}
	\caption{A sample video-grounded dialogue with a demonstration of a reasoning process}
	\label{fig:samples}
	\vspace{-0.1in}
\end{figure*}

Vision-language tasks have been studied to build intelligent systems that can perceive information from multiple modalities, such as images, videos, and text. 
Extended from imaged-grounded tasks, e.g. \cite{antol2015vqa}, recently \citet{jang2017tgif, lei-etal-2018-tvqa} propose to use video as the grounding features. 
This modification poses a significant challenge to previous image-based models with the additional temporal variance through video frames.  
Recently \citet{alamri2019audiovisual} further develop video-grounded language research into the dialogue domain.
In the proposed task, \emph{video-grounded dialogues}, the dialogue agent is required to answer questions about a video over multiple dialogue turns. 
Using Figure \ref{fig:samples} as an example, to answer questions correctly, a dialogue agent has to resolve references in dialogue context, e.g. ``he'' and ``it'', and identify the original entity, e.g. ``a boy" and ``a backpack". 
Besides, the agent also needs to identify the actions of these entities, e.g. ``carrying a backpack'' to retrieve information from the video. 

Current state-of-the-art approaches to video-grounded language tasks, e.g. \citet{le-etal-2019-multimodal, fan2019heterogeneous} have achieved remarkable performance through the use of deep neural networks to retrieve grounding video signals based on language inputs. 
However, these approaches often assume the reasoning structure, including resolving references of entities and detecting the corresponding actions to retrieve visual cues, is implicitly learned. 
An explicit reasoning structure becomes more beneficial as the tasks 
complicate in two scenarios: video with complex spatial and temporal dynamics, 
and language inputs with sophisticated semantic dependencies, e.g. questions positioned in a dialogue context. 
These scenarios often challenge researchers to interpret model hidden outputs, identify errors, and assess model reasoning capability.

Similar challenges have been observed in image-grounded tasks in which deep neural networks often exhibit shallow understanding capability as they simply exploit superficial visual cues \cite{agrawal-etal-2016-analyzing, goyal2017making, feng-etal-2018-pathologies, serrano-smith-2019-attention}. 
\citet{andreas2016neural} propose neural model networks (NMNs) by decomposing a question into sub-sequences called \emph{program} and assembling a network of neural operations.
Motivated by this line of research, we propose a new approach,
VGNMN, to video-grounded language tasks.
Our approach benefits from integrating neural networks with a compositional reasoning structure to exploit low-level information signals in video. An example of the reasoning structure can be seen on the right side of Figure \ref{fig:samples}.

Video-grounded Neural Module Network (VGNMN) tackles video understanding through action and entity-level NMNs to retrieve video features. 
We first decompose question into a set of entities and extract video features related to these entities. 
VGNMN then extracts the temporal steps by focusing on relevant actions that are associated with these entities.
VGNMN is analogous to human reasoning process by gradually retrieving linguistic and visual information from input modalities using a set of subjects and their actions. 

To tackle dialogue understanding, VGNMN is trained to resolve any co-reference in language inputs, e.g. questions in a dialogue context, to identify the unique entities in each dialogue.  
Previous approaches to video-grounded dialogues often obtain question global representations in relation to dialogue context.
These approaches might be suitable to represent general semantics in open-domain dialogues \cite{serban2016building} but they are not ideal to detect fine-grained entity-based information as the dialogue context evolves over several turns. 
However, in a video-grounded dialogue, it is empirical to dissect the dialogue context and detect the specific entity dependencies between questions and past dialogue turns. 

In summary, we introduce a neural module network approach to video-grounded language tasks  through a reasoning pipeline with entity and action representations applied to the spatio-temporal dynamics of video. 
In our evaluation, we achieve competitive performance on the large-scale benchmark Audio-visual Scene-aware Dialogues (AVSD) \cite{alamri2019audiovisual}.
Moreover, our experiments and ablation analysis indicate better model interpretability and robustness against the complexity in videos and dialogues. 
Finally, We adapt VGNMN for video QA and obtain significant performance on the TGIF-QA benchmark \cite{jang2017tgif} across all tasks.  

\section{Related Work}
Our work is related to three research areas: 
video-language understanding, video-grounded dialogues, and neural module networks. 

\subsection{Video-Language Understanding}
Video QA has been a proxy for evaluating a model's understanding capability of language and video and the task is treated as a visual information retrieval task. 
\citet{jang2017tgif, gao2018motion, jiangdivide} propose to learn attention guided by question global representation to retrieve spatial-level and temporal-level visual features. 
\citet{li2019beyond, fan2019heterogeneous, jiangreasoning} model interaction between all pairs of question token-level representations and temporal-level features of the input video through similarity matrix, memory networks, and graph networks respectively. 
\citet{gao2019structured, le2019learning, le2020hierarchical, leimulti, huang2020location} extends the previous approach by dividing a video into equal segments, sub-sampling video frames, or considering object-level representations of input video. 
We propose to replace token-level and global question representations with compositional question representations of specific entities and actions.
Visual information is retrieved as a sequential reasoning program over these entities and actions for more transparent and accurate information extraction. 

\subsection{Video-grounded Dialogues} 
Extended from video QA, video-grounded dialogue is an emerging task that combines dialogue response generation and video-language understanding research. 
As an orthogonal extension, this task entails a novel requirement for models to learn dialogue semantics and decode entity co-references in questions.
\citet{nguyen2018film, hori2019avsd, hori2019joint, sanabria2019cmu, le2019end, le-etal-2019-multimodal} extend traditional QA models by adding dialogue history neural encoders. 
\citet{kumar2019leveraging} enhances dialogue features with topic-level representations to express the general topic in each dialogue. 
\citet{schwartz2019factor} treats each dialogue turn as an independent sequence and allows interaction between questions and each dialogue turn. 
\citet{le-etal-2019-multimodal} encodes dialogue history as a sequence with embedding and positional representations. 
\citet{sanabria2019cmu} considers the task as a video summary task and concatenates question and dialogue history into a single sequence and proposes to transfer parameter weights from a large-scale video summary model. 
Different from prior work, we dissect the question sequence and explicitly detect and decode any entities and their references.
Our models also benefit from the insights on how models rely on the bias of component linguistic inputs to extract visual information. 

\subsection{Neural Module Network} 
Extending from research on neural semantic parsing \cite{jia-liang-2016-data, liang-etal-2017-neural}, \citet{andreas2016neural, andreas-etal-2016-learning} introduce NMNs to address visual QA by decomposing questions into linguistic sub-structures, known as programs, to instantiate a network of neural modules. 
NMN models have achieved significant success in synthetic image domains where a complex reasoning process is required \cite{johnson2017inferring, hu2018explainable, han2019visual}.
\citet{yi2018neural, han2019visual, mao2018the} improve previous work by decoupling visual-language understanding and visual concept learning. 
Our work is related to the recent work that extends NMN models to real data domains. 
For instance, \citet{kottur2018visual, jiang-bansal-2019-self, Gupta2020Neural} extend NMNs to visual dialogues and reading comprehension tasks.  
In this paper, we introduce a new approach that exploits NMN to learn dependencies between the composition in language inputs and the spatio-temporal dynamics in videos. 
This is not present in prior NMN models which are designed to apply on a two-dimensional image input without temporal variance. 
We propose to construct a reasoning structure from text, from which detected entities are used to extract visual information in the spatial space and detected actions are used to find visual information in the temporal space. 
\section{Method}
\subsection{Task Definition}
The input to the model consists of a dialogue $\mathcal{D}$ which is grounded on a video $\mathcal{V}$.
The input components include the question of current dialogue turn $\mathcal{Q}$, dialogue history $\mathcal{H}$, and the features of the input video, including visual and audio input. 
The output is a dialogue response, denoted as $\mathcal{R}$.
Each text input component is a sequence of words $w_1, ..., w_m \in \mathbb{V}^{in}$, the input vocabulary. 
Similarly, the output response $\mathcal{R}$ is a sequence of tokens $w_1, ..., w_n \in \mathbb{V}^{out}$, the output vocabulary. 
The objective of the task is the generation objective that output answers of the current dialogue turn $t$: 

\begin{align*}
    \hat{\mathcal{R}}_t &= \argmax_{\mathcal{R}_t} \displaystyle P(\mathcal{R}_t|\mathcal{V}, \mathcal{H}_t, \mathcal{Q}_t; \theta) \\
    &= \argmax_{\mathcal{R}_t} 
    \prod\limits_{n=1}^{L_\mathcal{R}}
    \displaystyle P_m(w_n| \mathcal{R}_{t,1:n-1},\mathcal{V}, \mathcal{H}_t, \mathcal{Q}_t; \theta) 
\end{align*}

In a Video-QA task, the dialogue history $\mathcal{H}$ is simply absent and the output response is typically collapsed to a single-token response.

\subsection{Encoders} 
\label{subsec:encoders}
\begin{table*}[htbp]
\resizebox{1.0\textwidth}{!} {
\begin{tabular}{llll}
\hline
\multicolumn{1}{c}{\textbf{Module}} & \multicolumn{1}{c}{\textbf{Input}} & \multicolumn{1}{c}{\textbf{Output}} & \multicolumn{1}{c}{\textbf{Description}}                                                   \\
\hline
\texttt{find}                                & \texttt{P}, \texttt{H}                          & $\texttt{H}_\mathrm{ent}$                                & For related entities in question, select the relevant tokens from dialogue history           \\
\texttt{summarize}                           & $\texttt{H}_\mathrm{ent}, \texttt{Q}$                            & $\texttt{Q}_\mathrm{ctx}$                                & Based on contextual entity representations, summarise the question semantics                  \\
\texttt{where}                               & $\texttt{P}, \texttt{V}$                            & $\texttt{V}_\mathrm{ent}$                                & Select the relevant spatial position corresponding to original (resolved) entities \\
\texttt{when}                                & $\texttt{P},\texttt{V}_\mathrm{ent}$                             & $\texttt{V}_\mathrm{ent+act}$                                & Select the relevant entity-aware temporal steps corresponding to the action parameter               \\
\texttt{describe}                            & $\texttt{P},\texttt{V}_\mathrm{ent+act}$                             & $\texttt{V}_\mathrm{ctx}$                                & Select visual entity-action features based on non-binary question types                        \\          
\texttt{exist}                            & $\texttt{Q},\texttt{V}_\mathrm{ent+act}$                             & $\texttt{V}_\mathrm{ctx}$                                & Select visual entity-action features based on binary (yes/no) question types                      \\ 
\hline
\end{tabular}
}
\caption{Description of the modules and their functionalities. We denote $P$ as the parameter to instantiate each module, $H$ as the dialogue history, $Q$ as the question of the current dialogue turn, and $V$ as video input.}
\label{tab:neural_modules}
\vspace{-0.2in}
\end{table*}

\textbf{Text Encoder.}
A text encoder is shared to encode text inputs, including dialogue history, questions, and captions. 
The text encoder converts each text sequence $\mathcal{X}=w_1, ..., w_m$ into a sequence of embeddings ${X} \in \mathbb{R}^{m \times d}$.
We use a trainable embedding matrix to map token indices to vector representations of $d$ dimensions through a mapping function $\phi$.
These vectors are then integrated with ordering information of tokens through a positional encoding function with layer normalization \cite{ba2016layer, vaswani17attention}. 
The embedding and positional representations are combined through element-wise summation. 
The encoded dialogue history and question of the current turn are defined as $H=\mathrm{Norm}(\phi(\mathcal{H}) + \mathrm{PE}(\mathcal{H})) \in \mathbb{R}^{L_\mathrm{H} \times d}$ and $Q=\mathrm{Norm}(\phi(\mathcal{Q}) + \mathrm{PE}(\mathcal{Q})) \in \mathbb{R}^{L_\mathrm{Q} \times d}$.

To decode program and response sequences auto-repressively, a special token ``\emph{\_sos}'' is concatenated as the first token $w_0$. The decoded token $w_1$ is then appended to $w_0$ as input to decode $w_2$ and so on. 
Similarly to input source sequences, at decoding time step $j$, the input target sequence is encoded to obtain representations for dialogue understanding program $P_{\mathrm{dial}}|^{j-1}_0$, video understanding program ${P}_{\mathrm{vid}}|_0^{j-1}$, and system response ${R}|_0^{j-1}$.
We combine vocabulary of input and output sequences and share the embedding matrix $E \in \mathbb{R}^{|\mathbb{V}| \times d}$ where $\mathbb{V}=\mathbb{V}^{in} \cap \mathbb{V}^{out}$.

\textbf{Video Encoder.}
To encode video, we use pre-trained models to extract visual and audio features. We denote $F$ as the sampled video frames or video clips.
For object-level visual features, we denote $O$ as the maximum number of objects considered in each frame. The resulting output from a pretrained object detection model is $Z_\mathrm{obj} \in \mathbb{R}^{F \times O \times d_\mathrm{vis}}$.
We concatenate each object representation with the corresponding coordinates projected to $d_\mathrm{vis}$ dimensions. 
We also make use of a CNN-based pre-trained model to obtain features of temporal dimension $Z_\mathrm{cnn} \in \mathbb{R}^{F \times d_\mathrm{vis}}$.
The audio feature is obtained through a pretrained audio model, $Z_\mathrm{aud} \in \mathbb{R}^{F \times d_\mathrm{aud}}$. We passed all video features through a linear transformation layer with ReLU activation to the same embedding dimension $d$.
\subsection{Neural Modules}
\label{subsec:neural_modules}
\begin{figure*}[h]
	\centering
	\resizebox{1.0\textwidth}{!} {
	\includegraphics{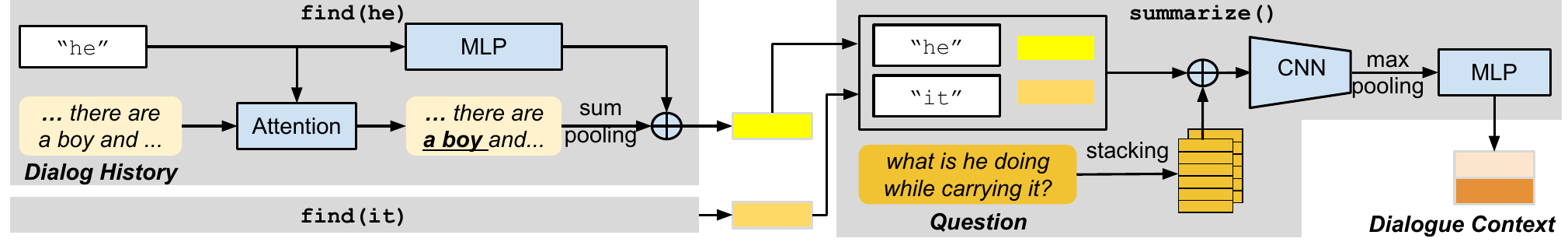}
	}
	\caption{\texttt{find} and \texttt{summarize} neural modules for dialogue understanding}
	\label{fig:dialog_modules}
	\vspace{-0.1in}
\end{figure*}

\begin{figure*}[htbp]
	\centering
	\resizebox{1.0\textwidth}{!} {
	\includegraphics{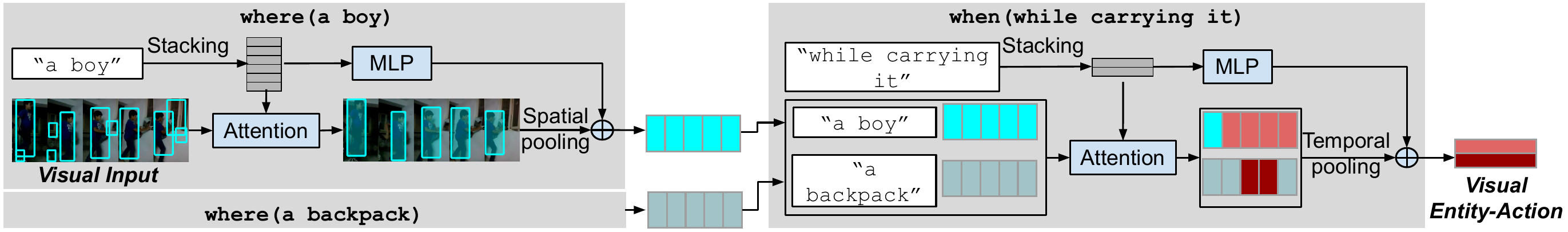}
	}
	\caption{\texttt{where} and \texttt{when} neural modules for video understanding}
	\label{fig:video_modules}
	\vspace{-0.1in}
\end{figure*}

We introduce neural modules that are used to assemble an executable program constructed by the generated sequence from question parsers. We provide an overview of neural modules in Table \ref{tab:neural_modules} and demonstrate dialogue understanding and video understanding modules in Figure \ref{fig:dialog_modules} and \ref{fig:video_modules} respectively. Each module parameter, e.g. ``a backpack'', is extracted from the parsed program (See Section \ref{subsec:question_parsers}). For each parameter, we denote $P \in \mathbb{R}^d$ as the average pooling of component token embeddings. 

\textbf{\texttt{find(P,H)$\rightarrow$H$_\mathrm{ent}$}}. 
This module handles entity tracing by obtaining a distribution over tokens in the dialogue history. 
We use an entity-to-dialogue-history attention mechanism applied from an entity $P_i$ to all tokens in the dialogue history. Any neural network that learn to generate attention between two tensors is applicable .e.g. \cite{DBLP:journals/corr/BahdanauCB14, vaswani17attention}.
The attention matrix normalized by softmax, $A_\mathrm{find,i} \in \mathbb{R}^{L_\mathrm{H}}$, is used to compute the weighted sum of dialogue history token representations. The output is combined with entity embedding $P_i$ to obtain contextual entity representation $H_\mathrm{ent,i} \in \mathbb{R}^d$.

\textbf{\texttt{summarize(H$_\mathrm{ent}$,Q)$\rightarrow$Q$_\mathrm{ctx}$}}. 
For each contextual entity representation $H_\mathrm{ent,i}$, $i=1,...,N_\mathrm{ent}$, it is projected to $L_\mathrm{Q}$ dimensions and is combined with question token embeddings through element-wise summation to obtain entity-aware question representation $Q_\mathrm{ent,i} \in \mathbb{R}^{L_\mathrm{Q} \times d}$. 
It is fed to a one-dimensional CNN with max-pooling layer \cite{kim-2014-convolutional} to obtain a contextual entity-aware question representation.  
We denote the final output as $Q_\mathrm{ctx} \in \mathbb{R}^{N_\mathrm{ent} \times d}$.

While previous models usually focus on global or token-level dependencies \cite{hori2019avsd, le-etal-2019-multimodal} to encode question features, our modules compress fine-grained question representations at the entity level. 
Specifically, \texttt{find} and \texttt{summarize} modules can generate entity-dependent local and global representations of question semantics. 
We show that our modularized approach can achieve better performance and transparency than traditional approaches to encode dialogue context \cite{serban2016building, vaswani17attention} (Section \ref{sec:results}). 

\textbf{\texttt{where(P,V)$\rightarrow$V$_\mathrm{ent}$}}.
Similar to the \texttt{find} module, this module handles entity-based attention to the video input. However, the entity representation $P$, in this case, is parameterized by the original entity in dialogue rather than in question (See Section \ref{subsec:question_parsers} for more description). 
Each entity $P_i$ is stacked to match the number of sampled video frames/clips $F$. An attention network is used to obtain entity-to-object attention matrix $A_\mathrm{where,i} \in \mathbb{R}^{F \times O}$. The attended feature are compressed through weighted sum pooling along the spatial dimension, resulting in $V_\mathrm{ent,i} \in \mathbb{R}^{F \times d}$, $i=1,...,N_\mathrm{ent}$. 

\textbf{\texttt{when(P,V$_\mathrm{ent}$)$\rightarrow$V$_\mathrm{ent+act}$}}.
This module follows a similar architecture as the \texttt{where} module. However, the action parameter $P_i$ is stacked to match $N_\mathrm{ent}$ dimensions. The attention matrix $A_\mathrm{when,i} \in \mathbb{R}^{F}$ is then used to compute the visual entity-action representations through weighted sum along the temporal dimension. We denote the output for all actions $P_i$ as $V_\mathrm{ent+act} \in \mathbb{R}^{N_\mathrm{ent} \times N_\mathrm{act} \times d}$

\textbf{\texttt{describe(P,V$_\mathrm{ent+act}$)$\rightarrow$V$_\mathrm{ctx}$}}. This module is a linear transformation to compute $V_\mathrm{ctx}={W_\mathrm{desc}}^{T}[V_\mathrm{ent+act}; P_\mathrm{stack}] \in \mathbb{R}^{N_\mathrm{ent} \times N_\mathrm{act} \times d}$ where $W_\mathrm{desc} \in \mathrm{R}^{2d \times d}$, $P_\mathrm{stack}$ is the stacked representations of parameter embedding $P$ to $N_\mathrm{ent} \times N_\mathrm{act}$ dimensions, and $[;]$ is the concatenation operation.

The \texttt{exist} module is a special case of \texttt{describe} module where the parameter $P$ is the average pooled question embeddings. 
The above \texttt{where} module is applied to object-level features. For temporal-based features such as CNN-based and audio features, the same neural operation is applied along the temporal dimension. Each resulting entity-aware output is then incorporated to frame-level features through element-wise summation.

An advantage of our architecture is that it separates dialogue and video understanding. 
We adopt a transparent approach to solve linguistic entity references during the dialogue understanding phase. 
The resolved entities are fed to the video understanding phase to learn entity-action dynamics in the video. 
We show that our approach is robust when dialogue evolves to many turns and video extends over time 
(Please refer to Section \ref{sec:results}).
\subsection{Question Parsers}
\label{subsec:question_parsers}
To learn compositional programs, we follow \cite{johnson2017clevr, hu2017learning} and consider program generation as a sequence-to-sequence task. 
We adopt a simple template ``$\langle\mathrm{param}_1\rangle\langle\mathrm{module}_1\rangle\langle\mathrm{param}_2\rangle\langle\mathrm{module}_2\rangle...$'' as the target sequence.
The resulting target sequences for dialogue and video understanding programs are sequences $\mathcal{P}_\mathrm{dial}$ and $\mathcal{P}_\mathrm{vid}$ respectively.

The parsers decompose questions into sub-sequences to construct compositional reasoning programs for dialogue and video understanding.
Each parser is a vanilla Transformer decoder, including multi-head attention layers on questions and past dialogue turns. 
For more details, please refer to the Supplementary Material. 
\subsection{Response Decoder}
System response is decoded by incorporating the dialogue context and video context outputs from the corresponding reasoning programs to target token representations. We follows a vanilla Transformer decoder architecture \cite{le-etal-2019-multimodal}, which consists of 3 attention layers: self-attention to attend on existing tokens, attention to $Q_\mathrm{ctx}$ from dialogue understanding program execution, and attention to $V_\mathrm{ctx}$ from video understanding program execution.
\begin{align*}
    A^{(1)}_\mathrm{res} &= \mathrm{Attention}(R|_0^{j-1},R|_0^{j-1},R|_0^{j-1}) \in \mathbb{R}^{j \times d}\\
    A^{(2)}_\mathrm{res} &= \mathrm{Attention}(A^{(1)}_\mathrm{res}, Q_\mathrm{ctx}, Q_\mathrm{ctx}) \in \mathbb{R}^{j \times d}\\
    A^{(3)}_\mathrm{res} &= \mathrm{Attention}(A^{(2)}_\mathrm{res}, V_\mathrm{ctx}, V_\mathrm{ctx}) \in \mathbb{R}^{j \times d}\\
\end{align*}
\textbf{Multimodal Fusion.}
For video features come from multiple modalities, visual and audio, the contextual features, denoted $V_\mathrm{ctx}$, is obtained through a weighted sum of component modalities, e.g. contextual visual features $V^\mathrm{vis}_\mathrm{ctx}$ and contextual audio features $V^\mathrm{aud}_\mathrm{ctx}$.
The scores $S_\mathrm{fusion}$ to compute the weighted sum is defined as: 
\begin{align*}
S_\mathrm{fusion} &= \mathrm{Softmax}(W_\mathrm{fusion}^T [Q_\mathrm{stack};V^\mathrm{vis}_\mathrm{ctx};V^\mathrm{aud}_\mathrm{ctx}])
\end{align*}
where $Q_\mathrm{stack}$ is the mean pooling output of question embeddings $Q$ which is then stacked to $N_\mathrm{ent} + N_\mathrm{act}$ dimensions, and $W_\mathrm{fusion} \in \mathbb{R}^{3d \times 2}$ are trainable model parameters.
The resulting $S_\mathrm{fusion}$ has a dimension of $\in \mathbb{R}^{(N_\mathrm{ent} + N_\mathrm{act}) \times 2}$.

\subsection{Optimization}
We optimize models by joint training to minimize the cross-entropy losses: 
\begin{align*}
   \mathcal{L} &= \alpha \mathcal{L}_\mathrm{dial} + \beta \mathcal{L}_\mathrm{vid} + \mathcal{L}_\mathrm{res} \\
    &= \alpha \sum_{j} -\log(\mathbf{P}_\mathrm{dial}(\mathcal{P}_\mathrm{dial,j})) \\
    &+ \beta \sum_{l} -\log(\mathbf{P}_\mathrm{video}(\mathcal{P}_\mathrm{video,l})) \\
    &+ \sum_{n} -\log(\mathbf{P}_\mathrm{res}(\mathcal{R}_\mathrm{n}))
\end{align*}

where $\mathbf{P}$ is the probability distribution of an output token. The probability is computed by passing 
output representations from the parsers and decoder 
to a linear layer $W \in \mathbb{R}^{d \times V}$ with softmax activation. We share the parameters between $W$ and embedding matrix $E$.

\section{Experiments}
\label{sec:results}
\subsection{Experimental Setup} 
We use the AVSD benchmark from the 7$^{th}$ Dialogue System Technology Challenge (DSTC7) \cite{hori2019avsd}.
The benchmark consists of dialogues grounded on the Charades videos \cite{sigurdsson2016hollywood}.
Each dialogue contains up to 10 dialogue turns, each turn consists of a question and expected response about a given video. 
For visual features, we use the 3D CNN-based features from a pretrained I3D model \cite{carreira2017quo} and object-level features from a pretrained FasterRNN model \cite{renNIPS15fasterrcnn}. 
The audio features are obtained from a pretrained VGGish model \cite{hershey2017cnn}.
In the experiments with AVSD, we consider two settings: one with video summary and one without video summary as input.
In the setting with video summary, the summary is concatenated to the dialogue history before the first dialogue turn. 
We also adapt VGNMN to the video QA benchmark TGIF-QA \cite{jang2017tgif}. 
For details of data and training procedure, please refer to the Supplementary. 
\begin{table*}[h]
\centering
\small
\resizebox{0.9\textwidth}{!} {
\begin{tabular}{lccccccc}
\hline 
\multicolumn{1}{c}{\textbf{Model}}                                                       & \textbf{PT} & \textbf{Visual} & \textbf{Audio} & \textbf{BLEU4} & \textbf{METEOR} & \textbf{ROUGE-L} & \textbf{CIDEr} \\\hline 
\multicolumn{7}{l}{\textbf{Audio/Visual only (without Video Summary/Caption)}}                                                                                                                                             \\
Baseline \cite{hori2019avsd}                                                                                    & -           & I               & -              & 0.305          & 0.217           & 0.481            & 0.733          \\
Baseline \cite{hori2019avsd}                                                                                     & -           & I               & V              & 0.309          & 0.215           & 0.487            & 0.746          \\
Baseline+GRU+Attn. \cite{le2019end}                                                               & -           & I               & V              & 0.315          & 0.239           & 0.509            & 0.848          \\
FGA \cite{schwartz2019factor}                                                                         & -                & I               & V              & -              & -               & -                & 0.806          \\
JMAN \cite{chu2020multi}                                                                         & -                & I               & -              & 0.309              & 0.240               & 0.520                & 0.890          \\
Student-Teacher \cite{hori2019joint}                                                                  &-                        & I               & V              & 0.371          & 0.248           & 0.527            & 0.966          \\
MTN \cite{le-etal-2019-multimodal}                                                                               &-                      & I               & -              & 0.343          & 0.247           & 0.520            & 0.936          \\
MTN \cite{le-etal-2019-multimodal}                                                                               &-                       & I               & V              & 0.368          & 0.259           & 0.537            & 0.964          \\
MSTN \cite{lee2020dstc8}                                                         &-                                 & I               & V              & 0.379          & 0.261           & 0.548            & 1.028          \\
BiST \cite{le-etal-2020-bist}                                                                    &                           & RX               & V              & 0.390          & 0.259           & \underline{0.552}            & 1.030          \\ 
RLM-GPT2 \cite{li2020bridging}                                                                    & \checkmark                          & I               & V              & \textbf{0.402}          & 0.254           & 0.544            & \underline{1.052}          \\ 
\hline
VGNMN                                                                          &-           & I               & -              & \underline{0.397}          & \underline{0.262}           & \textbf{0.550}            & \textbf{1.059}          \\
VGNMN                                                                          &-           & FR               & -              & 0.388         & 0.259           & 0.549            & 1.040          \\
VGNMN                                                                          &-           & -               & V              & 0.381         & 0.252         & 0.534            & 1.004          \\
VGNMN                                                                          & -          & I               & V              & 0.396          & \textbf{0.263}           & 0.549            & \textbf{1.059}          \\\hline
\multicolumn{7}{l}{\textbf{Audio/Visual only (with Video Summary/Caption)}}                                                                                                                                             \\
TopicEmb \cite{kumar2019leveraging}                                                                               &-                 & I               & A              & 0.329          & 0.223           & 0.488            & 0.762          \\
Baseline+GRU+Attn. \cite{le2019end}                                                     &-                      & I               & V              & 0.310          & 0.242           & 0.515            & 0.856          \\
JMAN \cite{chu2020multi}                                                                         & -                & I               & -              & 0.334              & 0.239               & 0.533                & 0.941          \\
FA+HRED \cite{nguyen2018film}                                                                   &-                              & I               & V              & 0.360          & 0.249           & 0.544            & 0.997          \\
VideoSum \cite{sanabria2019cmu}                                                   &-                                          & RX               & -              & 0.394          & 0.267           & 0.563            & 1.094          \\
VideoSum+How2 \cite{sanabria2019cmu}                                           &         \checkmark                                 & RX               & -              & 0.387          & 0.266           & 0.564            & 1.087          \\
MSTN \cite{lee2020dstc8}                                                         &-                                 & I               & V              & 0.377          & 0.275           & 0.566            & 1.115          \\
Student-Teacher \cite{hori2019joint}                                                         &-                                 & I               & V              & 0.405          & 0.273           & 0.566            & 1.118          \\
MTN \cite{le-etal-2019-multimodal}                                                            &-                                          & I               & -              & 0.392          & 0.269           & 0.559            & 1.066          \\
MTN \cite{le-etal-2019-multimodal}                                                              &-                                        & I               & V              & 0.410          & 0.274           & 0.569            & 1.129          \\ 
BiST \cite{le-etal-2020-bist}                                                                    &                           & RX               & V              & 0.429          & 0.284           & 0.581            & 1.192          \\ 
VGD-GPT2 \cite{le-hoi-2020-video}                                                                    & \checkmark                          & I               & V              & \underline{0.436}          & \underline{0.282}           & \underline{0.579}            & \underline{1.194}          \\ 
RLM-GPT2 \cite{li2020bridging}                                                                    & \checkmark                          & I               & V              & \textbf{0.459}          & \textbf{0.294}           & \textbf{0.606}            & \textbf{1.308}          \\ 
\hline
VGNMN             &-                                                                        & I               & -              & 0.421          & 0.277        & 0.574            & 1.171          \\
VGNMN             &-                                                                        & FR               & -              & 0.421           & 0.275           & 0.571           & 1.148          \\
VGNMN          &-                                                                            & I               & V              & 0.421          & 0.277           & 0.573            & 1.167          \\ 
VGNMN   &-     & I+C               & V              & {0.429}          & {0.278}           & {0.578}            & {1.188}          \\ \hline 
\end{tabular}
}
\caption{AVSD test results: The visual features are: I (I3D), ResNeXt-101 (RX), Faster-RCNN (FR), C (caption as a video input). The audio features are: VGGish (V), AclNet (A).
\checkmark on PT denotes models using pretrained weights and/or additional finetuning. The best and second-best results are bold and underlined respectively. 
}
\label{tab:avsd_results}
\vspace{-0.1in}
\end{table*}
\subsection{Experimental Results} 
\paragraph{AVSD Results.}
We evaluate model performance by the objective metrics, including BLEU \cite{papineni2002bleu}, METEOR \cite{banerjee2005meteor}, ROUGE-L \cite{lin2004rouge}, and CIDEr \cite{vedantam2015cider}, between each generated response and 6 reference gold responses.
As seen in  Table \ref{tab:avsd_results}, our models outperform most of existing approaches. 
In particular, the performance of our model in the setting without video summary input is comparable to the GPT-based RLM \cite{li2020bridging} with a much smaller model size. 
The Student-Teacher baseline \cite{hori2019joint} specifically focuses on the performance gap between models with and without textual signals from video summary through a dual network of expert and student models. 
Instead, VGNMN reduces this performance gap by efficiently extracting relevant visual/audio information based on fine-grained entity and action signals. 
We also found that VGNMN applied to object-level features is competitive to the model applied to CNN-based features. 
The flexibility of VGNMN neural programs show
when we treat the caption as an input equally to visual or audio inputs and execute entity-action level neural operations on the encoded caption sequence. 
\paragraph{Robustness.} 
To evaluate model robustness, we report BLEU4 and CIDEr of model variants in various experimental settings. 
Specifically, we compare against performance of output responses in the first dialogue turn position (i.e. 2$^{nd}$-10$^{th}$ turn vs. the 1$^{st}$ turn), or responses grounded on the shortest video length range (video ranges are intervals of 0-10$^{th}$, 10-20$^{th}$ percentile and so on).  
We report results of the following model variants: 
(1) \emph{w/o video NMN}: VGNMN without using video-based modules, e.g. \texttt{when} and \texttt{where}. Video features are retrieved through a token-level representation of questions \cite{le-etal-2019-multimodal}. 
(2) \emph{no NMN}: (1) + without dialogue-based modules, e.g. \texttt{find} and \texttt{summarize}. Dialogue history is encoded by a hierarchical LSTM encoder \cite{hori2019avsd}. 




In Table \ref{tab:robustness_video}, we compare VGNMN and model variant (1) to gain insights on model robustness to video complexity. 
We noted that the performance gap between VGNMN and (1) is quite distinct, with 7/10 cases of video ranges in which VGNMN outperforms.
However, in lower ranges (i.e. 1-23 seconds) and higher ranges (37-75 seconds), VGNMN performs not as well as model (1).  
We observed that related factors might affect the discrepancy, such as the complexity of the questions for these short and long-range videos. 
Potentially, our question parser for the video understanding program needs a more sophisticated decoding method (e.g. for tree-based program) to retrieve information in these ranges. 
\begin{table}[h]
\centering
\small
\resizebox{1.0\columnwidth}{!} {
\begin{tabular}{cllll}
\hline
                                                                                              & \multicolumn{2}{c}{\textbf{BLEU4}}                                          & \multicolumn{2}{c}{\textbf{CIDEr}}                                          \\
                                                                                             
\multicolumn{1}{c}{\textbf{\begin{tabular}[c]{@{}c@{}}video range \\ (seconds)\end{tabular}}} & \multicolumn{1}{c}{\textbf{VGNMN}} & \multicolumn{1}{c}{\textbf{Model (1)}} & \multicolumn{1}{c}{\textbf{VGNMN}} & \multicolumn{1}{c}{\textbf{Model (1)}} \\
\hline
1-23                                                                                          & 0.432                                & \textbf{0.447}                       & 1.298                                & \textbf{1.355}                       \\
23-28                                                                                         & \textbf{0.436}                       & 0.433                                & \textbf{1.264}                       & 1.165                                \\
28-30                                                                                         & \textbf{0.398}                       & 0.376                                & \textbf{1.203}                       & 1.164                                \\
30-30.6                                                                                       & \textbf{0.441}                       & 0.418                                & \textbf{1.220}                       & 1.202                                \\
30.6-31                                                                                       & \textbf{0.413}                       & 0.411                                & \textbf{1.250}                       & 1.166                                \\
31-31.6                                                                                       & 0.439                                & \textbf{0.451}                       & 1.249                                & \textbf{1.295}                       \\
31.6-32                                                                                       & \textbf{0.430}                       & 0.419                                & \textbf{1.217}                       & 1.192                                \\
32-33                                                                                         & \textbf{0.468}                       & 0.445                                & \textbf{1.343}                       & 1.237                                \\
33-37                                                                                         & \textbf{0.388}                       & 0.381                                & \textbf{1.149}                       & 1.124                                \\
37-75                                                                                         & 0.356                                & \textbf{0.365}                       & 0.910                                & \textbf{0.962}                      \\
\hline
\end{tabular}
}
\caption{Performance by video length between VGNMN variant (1) (VGNMN \emph{w/o video NMN}).}
\label{tab:robustness_video}
\vspace{-0.2in}
\end{table}
In Table \ref{tab:robustness_turn}, we gauge model robustness to the compelexity in dialogue. 
We observed that model (1) performs better than model (2) overall, especially in higher turn positions, i.e. from the 4$^{th}$ turn to 8$^{th}$ turn. Interestingly, we noted some mixed results in very low turn position, i.e. the 2$^{nd}$ and 3$^{rd}$ turn, and very high turn position, i.e. the 10$^{th}$ turn. Potentially, with a large dialogue turn position, the neural-based approach such as hierarchical RNN can better capture the global dependencies within dialogue context than the entity-based compositional NMN method. 
For additional analysis, please refer to the Supplementary. 
\begin{table}[h]
\centering
\small
\resizebox{1.0\columnwidth}{!} {
\begin{tabular}{cllll}
\hline
                                  & \multicolumn{2}{c}{\textbf{BLEU4}}                                          & \multicolumn{2}{c}{\textbf{CIDEr}}                                          \\
                                  
\multicolumn{1}{c}{\textbf{turn position}} & \multicolumn{1}{c}{\textbf{Model (1)}} & \multicolumn{1}{c}{\textbf{Model (2)}} & \multicolumn{1}{c}{\textbf{Model (1)}} & \multicolumn{1}{c}{\textbf{Model (2)}} \\
\hline
1                                 & 0.579                                & \textbf{0.587}                       & 1.623                                & \textbf{1.650}                       \\
2                                 & 0.429                                & \textbf{0.430}                       & \textbf{1.155}                       & 1.142                                \\
3                                 & 0.275                                & \textbf{0.289}                       & \textbf{0.867}                       & 0.846                                \\
4                                 & \textbf{0.309}                       & 0.305                                & \textbf{0.859}                       & 0.855                                \\
5                                 & \textbf{0.355}                       & 0.335                                & \textbf{1.088}                       & 1.023                                \\
6                                 & \textbf{0.357}                       & 0.329                                & \textbf{1.044}                       & 0.950                                \\
7                                 & \textbf{0.342}                       & 0.325                                & \textbf{0.896}                       & 0.847                                \\
8                                 & \textbf{0.361}                       & 0.332                                & \textbf{1.025}                       & 0.973                                \\
9                                 & 0.383                                & \textbf{0.431}                       & 1.043                                & \textbf{1.182}                       \\
10                                & \textbf{0.395}                       & 0.371                                & 0.931                                & \textbf{0.977}\\
\hline
\end{tabular}
}
\caption{Performance by dialogue turn between variants (1) (VGNMN \emph{w/o video NMN}) and (2) (\emph{no NMN})}
\label{tab:robustness_turn}
\vspace{-0.2in}
\end{table}
\begin{figure*}[h]
	\centering
	\resizebox{0.75\textwidth}{!} {
	\includegraphics{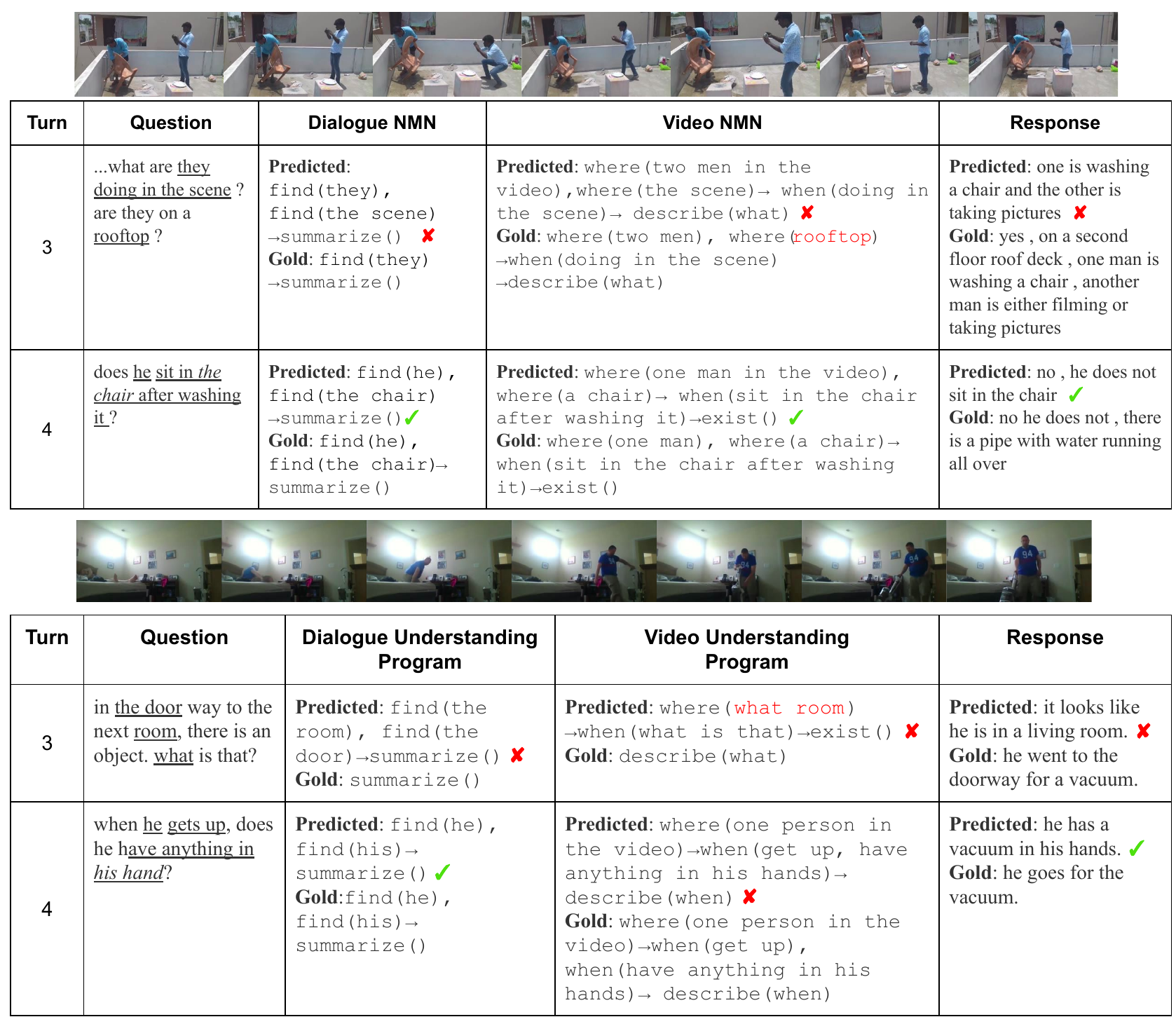}
	}
	\caption{Interpretability of model outputs from a dialogue in the test split of the AVSD benchmark. We demonstrate Example A (Top) and Example B (Bottom). 
	}
	\label{fig:intepretable}
	\vspace{-0.1in}
\end{figure*}
\paragraph{Interpretability.}
From Figure \ref{fig:intepretable}, we observe that in cases where predicted dialogue programs and video programs match or are close to the gold labels, the model can generate generally correct responses.  
For cases where some module parameters do not exactly match but are closed to the gold labels, the model can still generate responses with the correct visual information (e.g. the 4$^{th}$ turn in example B). 
In cases of wrong predicted responses, we can further look at how the model understands the questions based on predicted programs. 
In the 3$^{rd}$ turn of example A, the output response is missing a minor detail as compared to the label response because the video program fails to capture ``rooftop'' as a \texttt{where} parameter. 
These subtle yet important details can determine whether output responses can fully address user queries. 
In the 3$^{rd}$ turn of example B, the model wrongly identifies ``what room'' as a \texttt{where} parameter and subsequently generates a response to indicate it is ``a living room''.
For additional qualitative analysis, please refer to the Supplementary. 
\paragraph{TGIF-QA Results.} 
We report the result using the L2 loss in \emph{Count} task and accuracy in other tasks. 
From Table \ref{tab:tgifqa_results}, VGNMN outperforms the majority of the baseline models in all tasks by a large margin. 
Compared to AVSD experiments, the TGIF-QA experiments emphasize the video understanding ability of the models, removing the requirement for dialogue understanding and natural language generation.
Since TGIF-QA questions follow a very specific question type distribution (count, action, transition, and frameQA), the question structures are simpler and easier to learn than AVSD. Using exact-match accuracy of parsed programs vs. label programs as a metric, our question parser can achieve a performance 81\% to 94\% accuracy in TGIF-QA vs. 41-45\% in AVSD. 
The higher accuracy in decoding a reasoning structure translates to better adaptation between training and test time, resulting in higher performance gains. 

\begin{table}[htbp]
\centering
\small
\resizebox{1.0\columnwidth}{!} {
\begin{tabular}{llcccc}
\hline 
\multicolumn{1}{c}{\textbf{Model}}     & \multicolumn{1}{c}{\textbf{Visual}} & \textbf{\begin{tabular}[c]{@{}c@{}}Count \\ (Loss)\end{tabular}} & \textbf{\begin{tabular}[c]{@{}c@{}}Action \\ (Acc)\end{tabular}} & \textbf{\begin{tabular}[c]{@{}c@{}}Transition \\ (Acc)\end{tabular}} & \textbf{\begin{tabular}[c]{@{}c@{}}FrameQA\\  (Acc)\end{tabular}} \\
\hline 
VIS(avg) \cite{ren2015exploring}                              & R                                    & 4.80                  & 0.488                 & 0.348                     & 0.350                  \\
MCB (aggr) \cite{fukui-etal-2016-multimodal}                            & R                                    & 5.17                  & 0.589                 & 0.243                     & 0.257                  \\
Yu et al. \cite{yu2017end}                             & R                                    & 5.13                  & 0.561                 & 0.640                     & 0.396                  \\
ST-VQA (t) \cite{gao2018motion}                           & R+F                                  & 4.32                  & 0.629                 & 0.694                     & 0.495                  \\
Co-Mem \cite{gao2018motion}                                & R+F                                  & 4.10                  & 0.682                 & 0.743                     & 0.515                  \\
PSAC \cite{li2019beyond}                                  & R                                    & 4.27                  & 0.704                 & 0.769                     & 0.557                  \\
HME \cite{fan2019heterogeneous}                                    & R+C                                  & 4.02                  & 0.739                 & 0.778                     & 0.538                  \\
STA \cite{gao2019structured}                                   & R                                    & 4.25                  & 0.723                 & 0.790                     & 0.566                  \\
CRN+MAC \cite{le2019learning}                               & R                                    & 4.23                  & 0.713                 & 0.787                     & 0.592                  \\
MQL \cite{leimulti}                                   & V                                    & -                     & -                     & -                         & 0.598                  \\
QueST \cite{jiangdivide}                                & R                                    & 4.19                  & 0.759                 & 0.810                     & 0.597                  \\
HGA \cite{jiangreasoning}                                   & R+C                                  & 4.09                  & 0.754                 & 0.810                     & 0.551                  \\
GCN  \cite{huang2020location}                                  & R+C                                  & 3.95                  & 0.743                 & 0.811                     & 0.563                  \\
HCRN  \cite{le2020hierarchical}                                 & R+RX                                 & 3.82                  & 0.750                 & 0.814                     & 0.559                  \\  
BiST  \cite{le-etal-2020-bist}                                 & RX                                 & \textbf{2.14}                  & \textbf{0.847}                 & 0.819                     & 0.648                  \\
\hline
VGNMN & R                                    & 2.65                  & 0.845                 & \textbf{0.887}                     & \textbf{0.747}                  \\
\hline 
\end{tabular}
}
\caption{Experiment results on the TGIF-QA benchmark. The visual features are: ResNet-152 (R), C3D (C), Flow CNN from two-stream model (F), VGG (V), ResNeXt-101 (RX).    
}
\label{tab:tgifqa_results}
\vspace{-0.2in}
\end{table}

\bibliography{anthology,custom}
\bibliographystyle{acl_natbib}

\appendix

\section{Example Appendix}
\label{sec:appendix}

This is an appendix.

\end{document}